\documentclass[10pt,letterpaper]{article}

\usepackage{cogsci}
\cogscifinalcopy 

\usepackage{pslatex}
\usepackage{apacite}
\usepackage{natbib}
\usepackage{tipa}
\usepackage{float} 
\usepackage{amsmath}
\usepackage{graphicx}
\title{Phonetic accommodation and inhibition in a dynamic neural field model}
 
\author{{\large \bf Sam Kirkham (s.kirkham@lancaster.ac.uk)} \\
  Linguistics and English Language, Lancaster University, UK
  \AND {\large \bf Patrycja Strycharczuk (patrycja.strycharczuk@manchester.ac.uk)} \\
  Linguistics and English Language, University of Manchester, UK
  \AND {\large \bf Rob Davies (r.l.davies@lancaster.ac.uk)} \\
  Linguistics and English Language, Lancaster University, UK
  \AND {\large \bf Danielle Welburn (dani\_welburn@hotmail.com)} \\
  Linguistics and English Language, Lancaster University, UK
  }

\begin{document}

\maketitle

\begin{abstract}
Short-term phonetic accommodation is a fundamental driver behind accent change, but how does real-time input from another speaker's voice shape the speech planning representations of an interlocutor? We advance a computational model of change in speech planning representations during phonetic accommodation, grounded in dynamic neural field equations for movement planning and memory dynamics. A dual-layer planning/memory field predicts that convergence to a model talker on one trial can trigger divergence on subsequent trials, due to a delayed inhibitory effect in the more slowly evolving memory field. The model's predictions are compared with empirical patterns of accommodation from an experimental pilot study. We show that observed empirical phenomena may correspond to variation in the magnitude of inhibitory memory dynamics, which could reflect resistance to accommodation due to phonological and/or sociolinguistic pressures. We discuss the implications of these results for the relations between short-term phonetic accommodation and sound change.

\end{abstract}

\section{Introduction}

Spoken language is rarely static; when two people converse, they subtly (and sometimes not so subtly) modulate their voices in response to one another. This \textit{phonetic accommodation} is a fundamental characteristic of spoken interaction, representing the ebb and flow of human communication \citep{giles1973, pardo2006}. Experimental evidence for phonetic accommodation comes from the shadowing task paradigm, which involves a speaker reading a series of words, followed by imitating a (usually pre-recorded) model talker producing the same words \citep{goldinger1998}. This allows for the assessment of how much change occurs as a result of shadowing the model talker, while a post-test recording of the same words can be used to establish persistence of any adaptations. There is considerable evidence that speakers converge towards a model talker in the shadowing task and that listeners are also sensitive to these short-term changes \citep{goldinger1998, namy-etal2002, shockley-etal2004, tilsen2009b}. While accommodation is often cast as an automatic process \citep{goldinger1998}, \citet{babel2012} finds that accommodation is socially-mediated, with the degree of adaptation based on the perceived social characteristics of the model talker. 

Phonetic accommodation is short-term, but its accumulation over time is a key driver of accent change. Accent change generally slows significantly during adulthood, but change nonetheless can occur, especially with increased exposure to different accents \citep{evans-iverson2007, harrington-etal2019}. Exemplar models of speech processing hypothesize a mechanism behind these changes, such that the phonetic representations used in speech production are influenced by stored instances of language from speech perception \citep{gubian-etal2023, johnson2007, pierrehumbert2002}. In this sense, hearing another talker creates an episodic memory trace, which exerts a small bias on subsequent speech production. It stands to reason that over many such instances, small accent changes could occur, and this process of change would become accelerated if spread across a community.

In this study we advance a computational model of phonetic imitation, where imitation-based shadowing is used as an experimentally-constrained proxy for phonetic accommodation. Previous approaches are largely exemplar-based; e.g. \citet{goldinger1998} models data from a shadowing task using Minerva2 \citep{hintzman1984}. We take inspiration from this work, but advance an alternative dynamic neural field model \citep{schoener-etal2016}, which incorporates exemplar-like dynamics using a biophysically-inspired account of perception, action and memory \citep{gafos2006, tilsen2009b}. The motivation for this approach is that complex motor synergies are hypothesized to represent the locus of speech planning \citep{fowler1980, kelso-etal1986}, and dynamic field models are well-developed for auditory-motor dynamics underlying speech production and perception \citep{gafos2006, harper2021, roon-gafos2016, stern-shaw2023, tilsen2019}.

Our aim in this study is to examine excitatory and inhibitory dynamics in phonetic accommodation, inspired by previous work on response priming \citep{tilsen2007, roon-gafos2016}. We investigate how shadowing a model talker can lead to convergence on one trial, but propose a novel mechanism for observed divergence on subsequent trials. Rather than arising from online selective inhibition \citep{houghton-tipper1996}, divergence may reflect a delayed inhibitory effect in a coupled memory field, which we implement as a dual-layer model. As such, convergence during one trial can induce suppression in memory, leading to repulsion on subsequent trials even in the absence of an external input.

\section{Neural field model of phonetic accommodation}
\label{sec:model}

Our candidate model comes from the class of dynamic field models of movement planning \citep{erlhagen-schoener2002}, which are inspired by a long history of research in synergetics, self-organization and neural information processing \citep[e.g.][]{amari1977,grossberg1980, haken1977, kelso1995}. A dynamic neural field (DNF) functionally represents a neural population that is sensitive to a perceptual or movement parameter dimension. A DNF's evolution is shaped by inputs to the field, such as perceptual and task-related input, as well as memory dynamics and field interactions, such as self-excitation. DNFs are relatively well developed as models of the neural dynamics underpinning speech planning, execution and perception \citep[e.g.][]{gafos2006, kirkham-strycharczuk2024, roon-gafos2016, stern-shaw2023, tilsen2007, tilsen2019}, and see \citet{schoener-etal2016} for an introduction and different applications across the cognitive sciences.

\subsection{Model architecture}

We now outline a minimal model architecture for phonetic planning and memory during short-term phonetic accommodation. A dynamic neural field (DNF) evolves according to the \citet{amari1977} model that underpins Equation (\ref{dnf}):

\begin{multline}
\tau \dot{u}(x,t) = -u(x,t) + h + c_{\text{memory}} u_{\text{memory}}(x,t)
\\
+ c_{\text{auditory}} s_{\text{auditory}}(x,t) + c_{\text{response}}s_{\text{response}}(x,t)
\\
+ \int k(x-x') g(u(x',t))dx' + q\xi(x,t)
\label{dnf}
\end{multline}

where $\tau$ dictates the rate of field evolution, $-u(x,t)$ is time-dependent activation at each field site $x$, $h$ is the resting level of the neural field, $s(x,t)$ represents an input to the field, and $\xi(x,t)$ is Gaussian noise scaled by a coefficient $q$ \citep{schoener-etal2016}. As we are dealing with acoustic measurements, we assume that $x$ represents a one-dimensional reduction of the $F1\sim F2$ acoustic feature space, but a more realistic model could capture the coupling between acoustic, perceptual and articulatory representations using a multi-layer model.

Inputs $s(x,t)$ are Gaussian distributions over a parameter $x$ with amplitude $a$, centroid $p$ and width $w$,

\begin{equation}
s(x,t) = \sum_{i} a_{i} \exp \left[ - \frac{(x-p_{i})^2}{2w^2_{i}} \right].
\label{dnf_input}
\end{equation}

Response input $s_{\text{response}}(x,t)$ represents retrieval of a speech planning representation in response to the experiment's visual prompt and is weighted by $c_{\text{response}}$. We model this as retrieval of the appropriate representation from long-term memory \citep{roon-gafos2016}. Auditory input $s_{\text{auditory}}(x,t)$ is an auditory-perceptual input that couples the model talker's production to activation dynamics with strength $c_{\text{auditory}}$. We here treat $c_{\text{auditory}}$ as capturing the degree of attention to the incoming speech, which we expect is very high during a shadowing task, but lower in normal conversational interaction. In the shadowing block, auditory input cues the response input, whereas in the non-shadowing block the response is cued by a visual prompt (although we do not explicitly model the visual cue in this study).

The interaction kernel $k(x - x')$ in (\ref{kernel}) specifies excitatory and inhibitory forces across the activation field.

\begin{multline}
k(x - x') = \frac{c_{\text{excite}}}{\sqrt 2 \pi \sigma_{\text{excite}}} \exp \left[ -\frac{(x-x')^2}{2 \sigma^2_{\text{\text{excite}}}} \right]
\\
- \frac{c_{\text{inhibit}}}{\sqrt 2 \pi \sigma_{\text{inhibit}}} \exp \left[ -\frac{(x-x')^2}{2 \sigma^2_{\text{inhibit}}} \right] - c_{\text{global}}
\label{kernel}
\end{multline}

The interaction kernel is gated by a sigmoidal function

\begin{equation}
g(u) = \frac{1}{1+\exp (-\beta (u-\alpha))}
\label{dnf_sigmoid}
\end{equation}

where $\beta$ is the slope of the sigmoid and $\alpha$ is a threshold value of $u$. Each field location only contributes to above-threshold activation when it exceeds a threshold of $u = \alpha$, where typically $\alpha = 0$. Interaction generates local excitation and lateral inhibition, meaning that activation close to an input's centroid will be excited, whereas more distal activation will be inhibited. $c_{\text{excite}}$, $c_{\text{inhibit}}$ and $\sigma_{\text{excite}}$, $\sigma_{\text{inhibit}}$ are the mean and standard deviation of the excitatory/inhibitory components, and $c_{\text{global}}$ is a global inhibition constant.

The interaction kernel is a key part of our model, because any new inputs that are very close to the current activation peak will excite that location, causing activation to drift towards the input location, resulting in a compromise value between the speaker's planned target and the perceived target from the model talker \citep{erlhagen-schoener2002}. However, inhibitory forces mean that some inputs may cause dissimilation, causing activation to drift \textit{away} from the planned targets in a direction opposite the model talker's input \citep{tilsen2007}.

Short-term memory dynamics are achieved by a Hebbian layer \citep{samuelson-etal2011}. This is represented by the memory field $u_{\text{memory}}(x,t)$ in (\ref{eq:memory}) coupled to $u(x,t)$ with strength $c_{\text{memory}}$ and is subject to local interactions $w(x-x')$.

\begin{equation}
\dot{u}_{\text{mem}}(x,t)  =
\begin{cases} 
    \frac{1}{\tau_{\text{mem}}} [-u_{\text{mem}}(x,t) \\
    + \int w(x-x') g(u(x',t)) dx'], 
    & g(u) > \alpha \\[1em]
    \frac{1}{\tau_{\text{decay}}} [-u_{\text{mem}}(x,t)], 
    & g(u) \leq \alpha
\end{cases}
\label{eq:memory}
\end{equation}

When activation in the online planning field $u(x,t)$ is greater than threshold $\alpha$ the sigmoid $g(u)$ gates activation into the memory field at a rate determined by $\tau_{\text{mem}}$. When $g(u) \leq \alpha$ memory at those field locations undergoes decay at a rate of $\tau_{\text{decay}}$. The memory field evolves on a slower timescale than the field dynamics, while memory decay evolves the slowest, such that $ \tau_{\text{decay}} > \tau_{\text{mem}} > \tau$. This reflects the fact that memory formation happens faster than memory decay. This suggests that even if a speaker's current production is shifted in the direction of the model talker, the concomitant effects on the memory field will be relatively small. As such, we predict that any post-shadowing convergence or divergence will be minimal over a small number of trials.

Note that in our simulations we use an interaction kernel for both the main parameter field $k(x-x')$ and the memory field $w(x-x')$. The memory kernel is specified for local inhibition but not global inhibition, and we hypothesize that the memory field may have different interaction dynamics due to learning patterns and longer-term attentional dynamics. For example, previous research suggests that phonetic accommodation varies between different vowels \citep{evans-iverson2007, babel2012}. While in some cases this can be partly explained due to the distance between a field's current activation pattern and the model talker's input, there are cases where speakers converge during shadowing and diverge post-shadowing, which varies between vowels and speakers. We hypothesize that this could represent different patterns of lateral inhibition in \textit{memory} for different vowels, which may be a consequence of phonological, perceptual or sociolinguistic pressures. Note that we locate any such differences in the memory field and not in the online planning field. This predicts that speakers will typically converge towards the model talker, but may vary in the post-shadowing response depending on the current state of the memory field.

\subsection{Simulating experimental trials}

A simulated interaction proceeds as follows in three blocks.

\begin{enumerate}
	\item \textbf{Baseline}. The visual prompt cues $s_{\text{response}}(x,t)$ input, which raises activation above threshold and triggers production at the parameter value corresponding to peak activation. We assume $s_{\text{auditory}}(x,t) = 0$ during the baseline block, meaning it has no influence on activation. The field dynamics leave an activation trace in $u_{\text{memory}}(x,t)$.
	\item \textbf{Shadowing}. Auditory input from the model talker $s_{\text{auditory}}(x,t) > 0$ raises sub-threshold activation in $u(x,t)$ and the response input $s_{\text{response}}(x,t)$ subsequently raises activation above threshold and production occurs. High attention to the model talker reflected in a large $c_{\text{auditory}}$ value means that input amplitude is high, which causes its effects to persist over time. These dynamics leave an activation trace in the updated $u_{\text{memory}}(x,t)$ field.
	\item \textbf{Post-shadowing}. The visual prompt cues $s_{\text{response}}(x,t)$, while $s_{\text{auditory}}(x,t)$ = 0, which raises activation above threshold, cues production, and leaves a memory trace.
\end{enumerate}

Note that the memory field $u_{\text{memory}}(x,t)$ is active on each trial and shapes the evolution of the planning field based on its current state, which is also updated during each trial.

All simulations were implemented in Python using NumPy \citep{harris2020} and SciPy \citep{SciPy-NMeth2020}, with visualizations made using Matplotlib \citep{hunter2007}. Numerical solutions were calculated using an Explicit Runge-Kutta method of order 5(4) via SciPy's integrate.solve\_ivp function.

\section{A pilot experiment on phonetic accommodation}
\label{sec:experiment}

\subsection{Experiment design}

We also report an experimental pilot study of phonetic accommodation in a shadowing task, where speakers of Northern Anglo British English shadow a model talker with an accent different from their own. Due to the small sample size, our experiment is not an empirical assessment of the facts behind phonetic accommodation, but is instead used to generate plausible empirical scenarios. We subsequently attempt to model these empirical scenarios in order to generate hypothesized mechanisms behind patterns of accommodation.

The experiment featured three blocks: pre-test, shadowing, post-test \citep{babel2010, goldinger1998}. In the pre-test and post-test blocks, speakers read aloud single hVd words along with a set of target words. The shadowing block required speakers to identify and repeat the same target words produced by a model talker, who was a male Standard Southern British English speaker aged 21. We specifically focus on two vowels that differ substantially between the participants and model talker: \textsc{bath} and \textsc{strut}. Standard Southern British English realizes these vowels as \textipa{[A]} and \textipa{[2]} respectively, while in almost all varieties of Northern Anglo English these vowels are produced as \textipa{[a]} and \textipa{[U]}. These vowels represent the most characteristic difference between northern and southern varieties in England \citep{wells1982}; they are highly salient to listeners and can also undergo change as a consequence of long-term different-accent exposure \citep{evans-iverson2007}. The \textsc{bath} target words were \textit{bath, chance, fast, mast, staff}, while the \textsc{strut} words were \textit{strut, bust, chuck, fun, mud}. Another 10 non-\textsc{bath/strut} words were also included in all experimental blocks as distractor stimuli.

\begin{figure*}[ht]
\centering
\includegraphics[width=\textwidth]{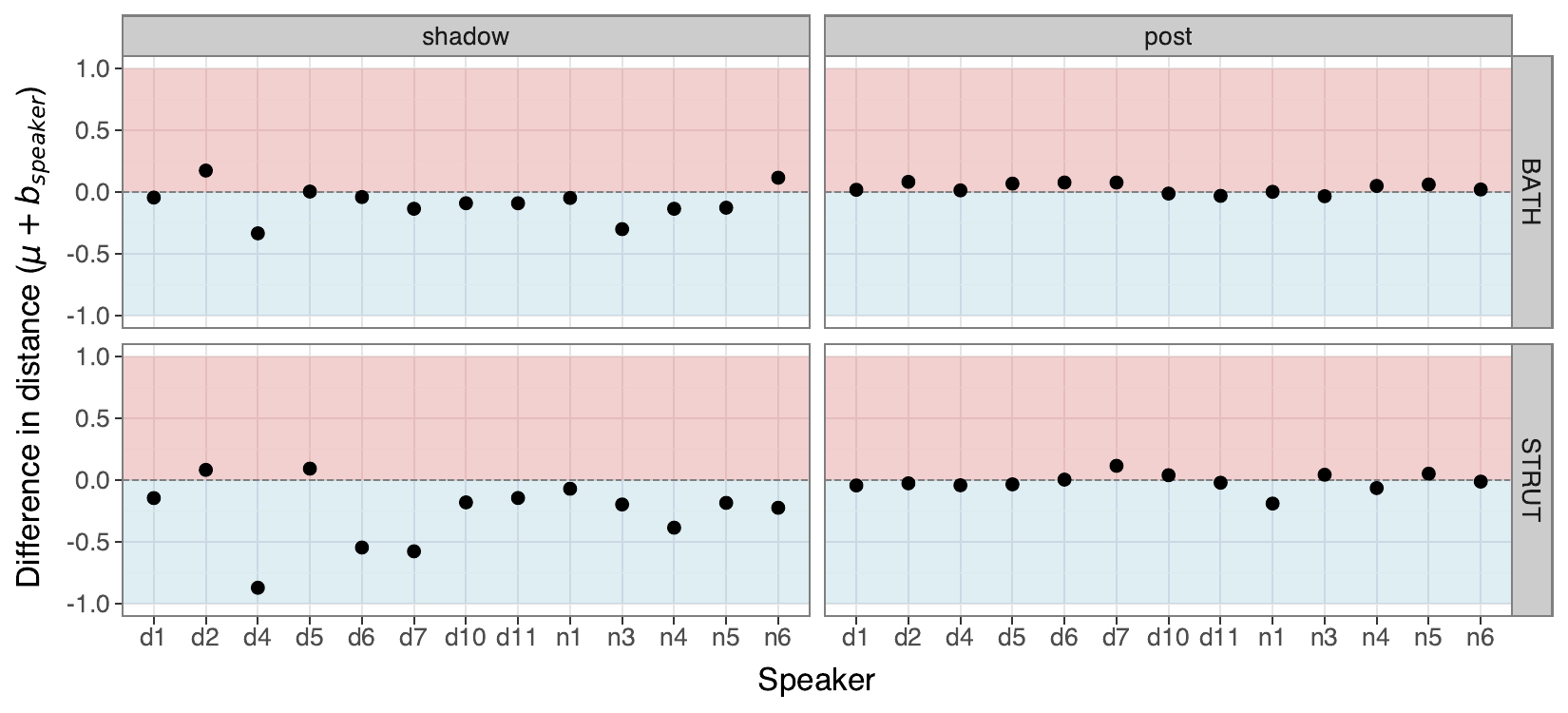}
\caption{By-speaker difference in distance values for vowel and block. Values are from the Bayesian model and represent each speaker's random intercept coefficient added to the model's grand intercept. Negative values indicate convergence towards the model talker (blue shading); positive values indicate divergence (red shading); zero values indicate no accommodation. Speaker labels (d/n) refer to data collected by different experimenters and do not reflect any differences in speaker characteristics.}
\label{fig:north_distances}
\end{figure*}

\subsection{Participants and recording}

All participants were first-language speakers of Northern Anglo English, aged between 19--22 years old. 18 speakers completed the experiment, with 13 speakers (11 female, 2 male) included in the analysis (two speakers were removed as they recognized the model talker, and three speakers were removed due to significant distortion in the audio recordings). The experiment was adminstered using PsychoPy \citep{psychopy2019}, with audio recorded in a sound-attenuated booth at 44.1 kHz using a Beyerdynamic Opus 55 headset microphone (5cm from the mouth), pre-amplified and digitized using a Sound Devices USBPre 2 audio interface. Audio stimuli were delivered using Beyerdynamic DT 770 headphones.

\subsection{Data processing}

Recordings were force-aligned using Montreal Forced Aligner \citep{mcauliffe-etal2017} and formant estimation was optimized using the FastTrack algorithm \citep{barreda2021, fruehwald_fasttrackpy}, with a 20-step search window of 4000--7000 Hz, 25 ms window length, 2 ms step size, 5th-order DCT smoothing. Formant values were extracted from vowel midpoints and by-speaker $z$-scored across hVd and target words. The degree of accommodation was quantified by calculating the Euclidean distance $d$ between each speaker's production and the corresponding model talker production.

We subtract the Euclidean distance value for the baseline block from the shadowing and post-test blocks in order to calculate a `difference in distance' metric that captures the degree of accommodation \citep{babel2012}. Values of zero indicate no accommodation, negative values indicate convergence (reduced distance from model talker), and positive values indicate divergence (increased distance from model talker). All data analysis was carried out in the Python programming language, using the packages NumPy \citep{harris2020}, Pandas \citep{pandas} and plotnine \citep{plotnine}.

\subsection{Results}

We fit four Bayesian linear mixed models to the difference in distance ($d$) values for each combination of task (shadowing/post) and vowel (\textsc{bath/strut}). Each model contains a grand intercept $\mu$ and random intercepts $b_{\text{word}}$ and $b_{\text{speaker}}$. Models were written in Stan and run using cmdstanpy \citep{stan2024} with weakly informative priors $\mu \sim \mathcal{N}(0, 2)$ and $b \sim \mathcal{N}(0, 1)$, using 4 MCMC chains, 500 warmup iterations, and 2000 sampling iterations. Speakers accommodate to the model talker during shadowing to a greater extent for \textsc{strut} ($\bar{d} = -0.25$, 95\% CI [$-$0.60, 0.13]) than \textsc{bath} ($\bar{d} = -0.08$, 95\% CI [$-$0.30, 0.15]). The vowels differ only minimally post-shadowing, with \textsc{bath} slightly diverging from the model talker post-shadowing ($\bar{d} = 0.03$, 95\% CI [$-$0.22, 0.25]), while \textsc{strut} is closer to the pre-shadowing baseline ($\bar{d} = 0.01$, 95\% CI [$-$0.19, 0.16]). Note that in all cases the wide credible intervals point towards extensive between-speaker variation and do not support vowel-specific differences on a group level.

Figure \ref{fig:north_distances} shows speaker-specific difference in distance values. These values represent the fitted by-speaker random intercepts, added to the model's grand intercept, which provides an estimate of each speaker's difference in distance value. The plot reflects some convergence in \textsc{bath}, with the majority of speakers below the zero line, followed by the majority of speakers showing a small amount of divergence post-shadowing. However, some speakers do show small divergence during shadowing, such as d2 and n6, but all speakers are very close to baseline or above post-shadowing. The \textsc{strut} data is more variable during shadowing, with some speakers showing substantial convergence to the model talker (e.g. d4, d6, d7). Two speakers diverge during shadowing for \textsc{strut} (d2, d5), one of whom also diverged for \textsc{bath} (d2). The post-shadowing \textsc{strut} data shows strong clustering around the zero line, indicating a return to the baseline production. Speaker n1 is the only one who is slightly closer to the model talker post-shadowing than during shadowing, but these differences remain small.

\subsection{Summary and next steps}

The overall picture from this sample is considerable variability in accommodation. Both vowels show minor amounts of divergence in post-shadowing, with \textsc{bath} showing slightly greater divergence on a speaker-specific level. While our results are insufficient to claim a robust vowel-specific effect, \citet{evans-iverson2007} report a longitudinal study in which northern speakers converge to SSBE \textsc{strut} to a greater extent than \textsc{bath} over a period of two years, which the authors suggest is due to the greater salience of \textsc{bath} in northern English. While our data do not support a group-level vowel effect, there are certainly individual speakers who follow this pattern. This points to a more generic observation that convergence to a model talker can potentially result in a subsequent return to baseline \textit{or} subsequent divergence. In the following section, we use our computational model to explore the mechanisms that could generate these two phenomena.

\section{Simulating interactional scenarios}
\subsection{Motivations and approach}
\label{sec:simulations}

We now use our model to replicate two potential observations: (1) convergence followed by return to baseline; (2) convergence followed by divergence. Our primary interest is identifying which mechanisms in our model are required to capture these patterns. We refer to these cases as \textsc{strut} (return to baseline) and \textsc{bath} (divergence) as this is the trend in the literature, but these should be taken as more general examples that are within the model's scope. To provide some empirical validity to the simulations, we focus on modelling the small vowel-specific effects from the empirical data. While these are very small, we view this as preferable to modelling potentially larger effects that are not observed in our data and may therefore be unrealistic.

For all simulations we centre the idealized speaker's memory trace at zero across a field of $x\in [-10, +10$] (the majority of the field either side of zero is subject to significant inhibition, so it is unlikely that such areas receive any significant activation). The inputs $s_{\text{auditory}}(x,t)$ are based on the average $z$-scored distance from the model talker, with \textsc{strut} $= -1.4$ and \textsc{bath} $= -1.2$. These input values represent differences from the idealized speaker's existing representation. The planning field interaction kernel $k(x-x')$ is defined as $c_{\text{excite}} = 2, \sigma_{\text{excite}} = 0.2, c_{\text{inhibit}} = 1, \sigma_{\text{inhibit}} = 2, c_{\text{global}} = 0.5$. The default memory kernel $w(x-x')$ is identical to the field kernel, except $c_{\text{global}} = 0$ and $\sigma_{\text{excite}} = 0.1$. Temporal parameters are $\tau = 25$, $\tau_{\text{memory}} = 150$, $\tau_{\text{decay}} = 500$, while $c_{\text{memory}} = 10$, $c_{\text{auditory}} = 10$, $c_{\text{response}} = 1$, $h = -2$, $q = 3$, $\beta=1.5$. All inputs $s(x,t)$ have $a = 10$ and $w = 0.5$.

All simulations lasted for a duration of 300 ms. Inputs are constant over time because we make the assumption that monophthongs are one-target vowels \citep{strycharczuk-etal2024}. To represent the acoustic parameter selected for speech production, we sample at the time-step corresponding to peak activation, with the $x$-location of peak activation representing the selected parameter value for production.

We first initialize short-term memory $u_{\text{memory}}(x,t)$ as a zero-valued flat field and then run simulations with a single input $a =100, p=0, w=0.5$, which when coupled to the memory field serves to update $u_{\text{memory}}(x,t)$ based on the resulting field activation. This represents the existing short-term memory for each vowel, while the response input $s_{\text{response}}(x,t)$ is drawn from longer-term phonological memory. For each simulation, the initial memory state is the memory state at the end of the previous simulation.

\subsection{Phonetic convergence and return to baseline}

The model straightforwardly captures the dynamics of phonetic convergence followed by return to near-baseline. In the \textsc{strut} vowel simulations, peak activation is at $x = - 0.22$ during shadowing and $x= 0.02$ during post-shadowing. This is close to the empirical means of $\bar{d} = -0.25$ and $\bar{d} = 0.01$, with a relative difference between shadowing and post-shadowing of $\bar{d}_{\text{diff}} = 0.26$ and $ x_{\text{diff}} = 0.24$, which represents good agreement between model and data.

\begin{figure}[ht]
\centering
\includegraphics[scale=0.35]{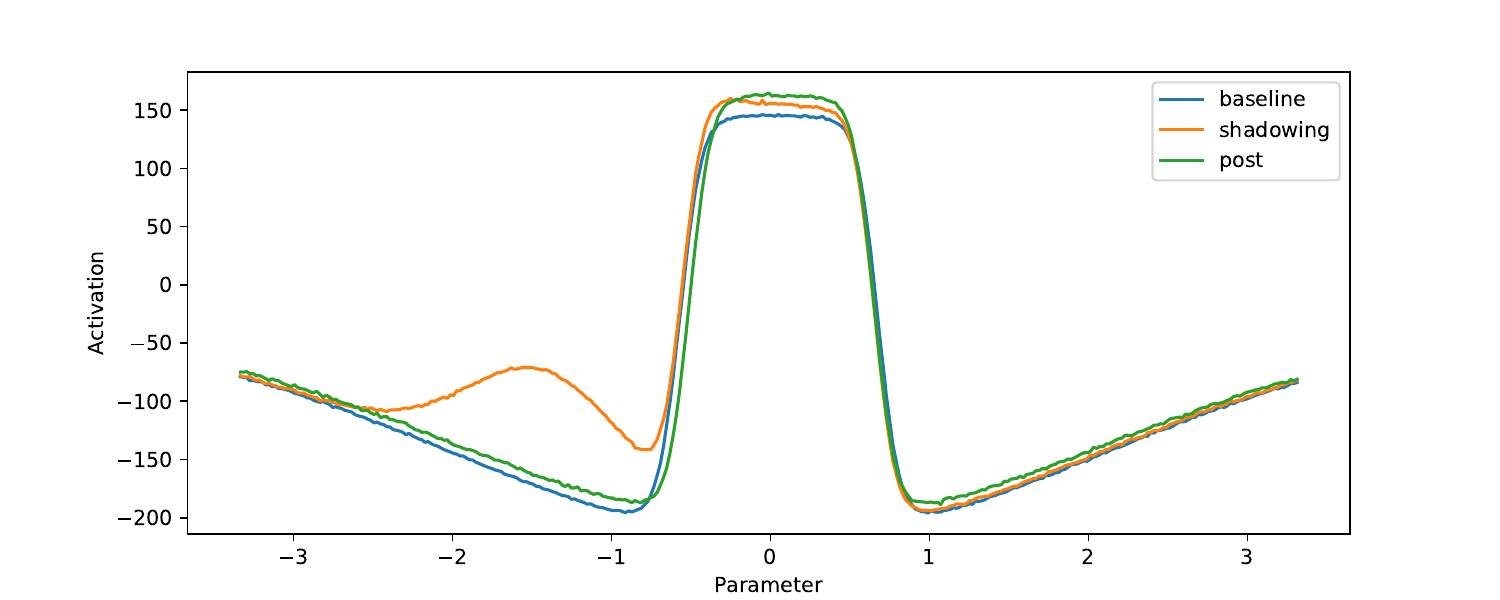}
\includegraphics[scale=0.35]{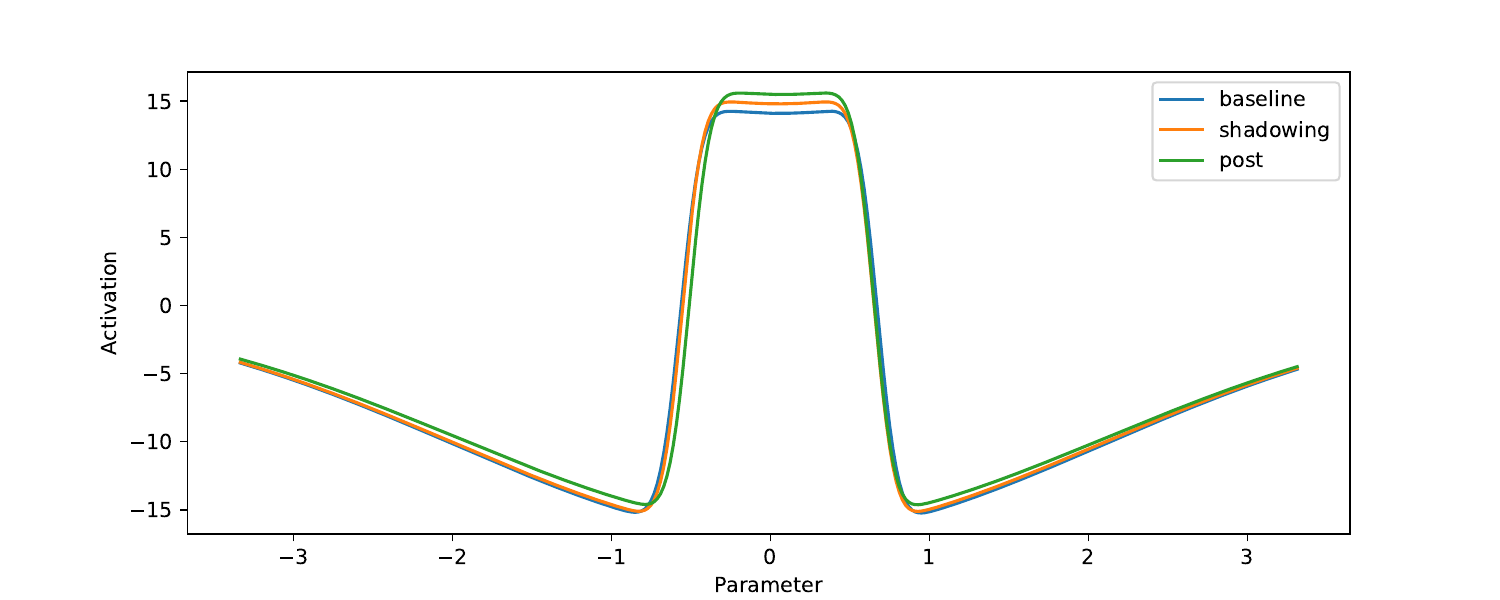}
\caption{Activation field (top) and memory field (bottom) for \textsc{strut} simulations. The $x$ parameter range has been truncated to highlight small differences in the activation peaks.}
\label{fig:dnf-strut}
\end{figure}

The occurrence of accommodation during shadowing versus return to (near) baseline post-shadowing is a consequence of the auditory input and inhibitory dynamics.  This is shown in Figure \ref{fig:dnf-strut} (top), where the small bump at the left of the activation field during shadowing (orange line) represents the effects of $s_{\text{auditory}}(x,t)$. While this input does not reach the activation threshold of 0, it does slightly pull the activation centroid leftwards, resulting in the small degree of observed accommodation towards the model talker. The degree of accommodation is attenuated as the input occurs in a region of parameter space that is subject to considerable inhibition (i.e. the negative values near the base of the primary activation peak). This small amount of accommodation has a very minimal effect on the memory field in Figure \ref{fig:dnf-strut} (bottom), with an almost undetectable rightwards shift of the memory peak as a consequence of greater inhibition in short-term memory. The memory field evolves more slowly than the parameter activation field, meaning that any production effects are very gradual. Note that while these effects are very small, they reflect the average empirical changes in speech production as a consequence of minimal short-term exposure to the model talker.

\subsection{Phonetic convergence followed by divergence}

We now examine how our model can reproduce the general effect of convergence (shadowing) followed by divergence (post-shadowing). We specifically model the small average effect for \textsc{bath}, but note that some speakers diverge to a much greater extent than this. In doing so, we turn to the memory kernel, which situates the differences between \textsc{bath} and \textsc{strut} in vowel-specific memories, rather than purely metric parameter differences. We run the same simulation as for \textsc{strut} but with three changes. First, $s_{\text{auditory}}(x,t)$ has $p = -1.2$ rather than $p = -1.4$ to reflect the empirical baseline distance from the model talker for \textsc{bath}. Second, the memory kernel has higher local inhibition, with $c_{\text{inhibit}} = 1.8$ (compared to $c_{\text{inhibit}} = 1$ for \textsc{strut}). Third, the memory kernel also has higher $\sigma_{\text{inhibit}} = 3$ (compared with $\sigma_{\text{inhibit}} = 2$ for \textsc{strut}). This specifies the memory kernel for \textsc{bath} as having stronger/wider local inhibition, which corresponds to the memory field's increased resistance in this region. This is in line with previous literature showing that \textsc{bath} is more resistant to accommodation than \textsc{strut} for northern speakers.

\begin{figure}[ht]
\centering
\includegraphics[scale=0.35]{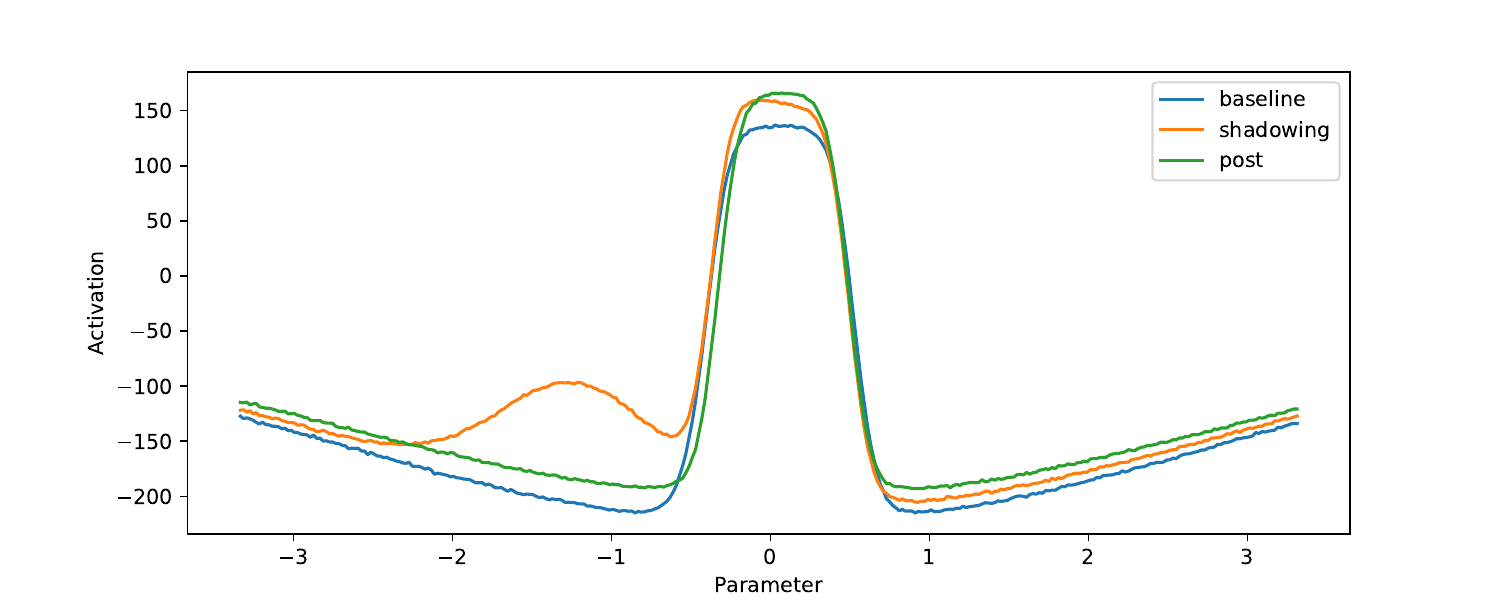}
\includegraphics[scale=0.35]{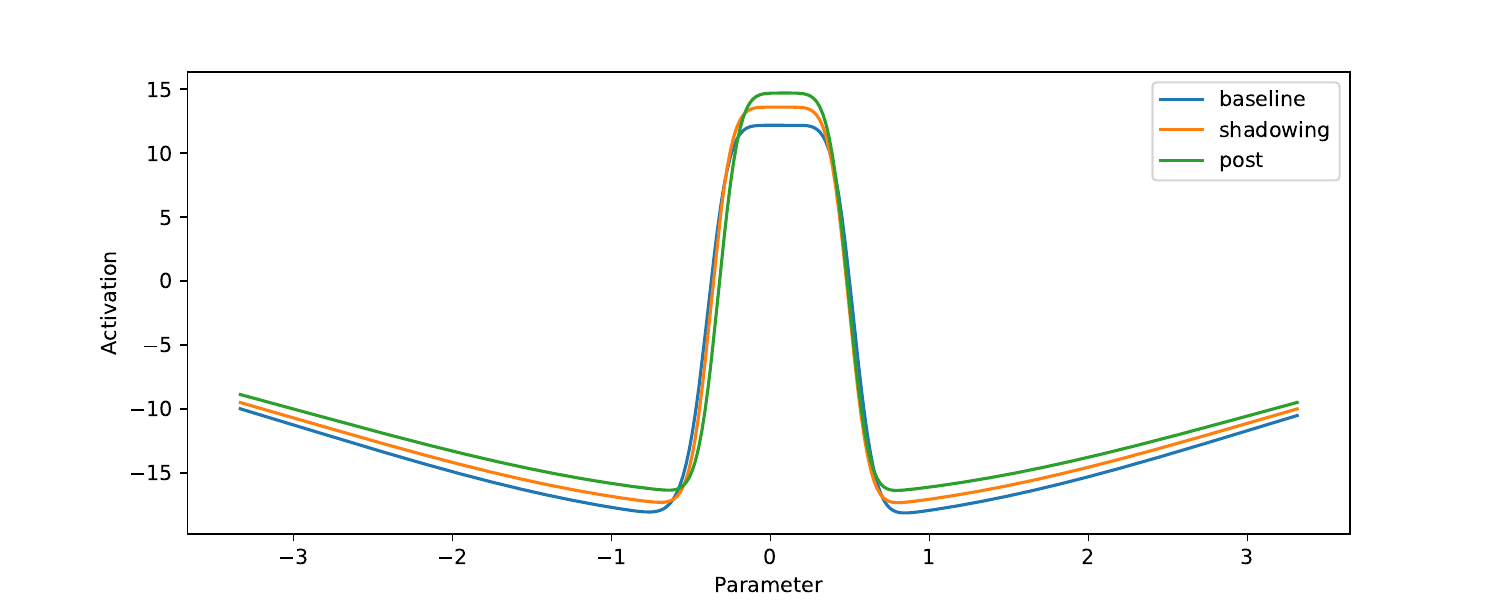}
\caption{Activation field (top) and memory field (bottom) for \textsc{bath} simulations. The $x$ parameter range has been truncated to highlight small differences in the activation peaks.}
\label{fig:dnf-bath}
\end{figure}

The resulting simulations for \textsc{bath} show peak activation at $x = - 0.1$ for shadowing and $x = 0.02$ for post-shadowing, which is close to the mean empirical magnitude of accommodation ($\bar{d} = -0.08$) and divergence ($\bar{d} = 0.03$). If we take the relative difference between shadowing and post-shadowing then $\bar{d}_{\text{diff}} = 0.11$ and $ x_{\text{diff}} = 0.12$, showing good agreement between model and data. This can be seen in Figure \ref{fig:dnf-bath}, where sub-threshold auditory input peak on the left-hand side (orange) pulls the activation peak slightly leftwards, representing accommodation, but inhibitory memory dynamics repel this slightly and the subsequent post-shadowing production is shifted slightly rightwards, representing divergence. Note that the differences between conditions in the memory trace are slightly larger than those in Figure \ref{fig:dnf-strut}, as a consequence of stronger inhibitory dynamics in the memory field for \textsc{bath}.

\section{Discussion}
\label{sec:discussion}

We proposed a dynamic neural field model of phonetic accommodation that qualitatively captures some observed empirical phenomena. While the majority of speakers show some convergence and a return to baseline, some speakers converge during shadowing and then diverge post-shadowing, resulting in a greater distance from the model talker \textit{after} shadowing. This cannot be straightforwardly modelled using a single-layer field, which leads us to propose a delayed inhibitory effect in memory, due to the different temporal scales of online planning and short-term memory. Indeed, previous resarch suggests that inhibition may be part of the long-term memory of a gesture \citep{tilsen2007} and we show that modelling these vowel differences in the memory field, rather than the planning field, exposes a potential mechanism.

Previous research finds vowel-specific differences in accommodation; while this can be explained by inhibitory differences, why should inhibitory dynamics vary between vowels?  One rationale is that changes in \textsc{bath} have structural implications for northern speakers, whose vowel in the \textsc{palm/start} lexical sets is phonetically similar to SSBE \textsc{bath}. Convergence would lead to potential merger between vowel categories, so greater inhibition may prevent category merger. Additionally, the \textsc{bath} vowel is a strong shibboleth of the north/south divide \citep{wells1982} and northern speakers tend to resist change in this vowel \citep{evans-iverson2007}. Our empirical data shows that inhibitory dynamics are not sufficient to completely block accommodation, suggesting an automatic dimension to accommodation \citep{goldinger1998}. However, inhibitory dynamics attenuate the magnitude of accommodation and trigger divergence in short-term memory, which bolsters the maintenance of this salient accent feature.

In the present study, we have only modelled the very small average effects from the statistical model, but varying degrees of accommodation and persistence can be modelled through variation in the input weighting (i.e. reflecting attention to the model talker) and in excitatory/inhibitory forces. Variation in these parameters is likely to be a potential locus of speaker-specific variation, which could explain differences in adaptation. For example, reducing memory inhibition for \textsc{bath} produces a return closer to baseline, rather than dissimilation.

In summary, vowel-specific phonetic accommodation can be modelled as differences in short-term inhibitory memory dynamics, which we hypothesize is motivated by phonological contrast and socially-motivated resistance to change. Vowels with weaker inhibitory dynamics are predicted to undergo greater accommodation, which if repeated over many interactions could lead to sound change. In future work we plan to conduct a more comprehensive experimental study, as well as integrate acoustic-perceptual representations with nonlinear gestural models of articulatory control \citep{kirkham2025b, kirkham2025, sorensen-gafos2016, stern-shaw2024}.

\clearpage
\section{Acknowledgments}

This research was funded by UKRI/AHRC grant AH/Y002822/1 awarded to SK.

\bibliographystyle{apacite}

\setlength{\bibleftmargin}{.125in}
\setlength{\bibindent}{-\bibleftmargin}

\bibliography{bibliography.bib}

\begin{thebibliography}{}

\bibitem [\protect \citeauthoryear {%
Amari%
}{%
Amari%
}{%
{\protect \APACyear {1977}}%
}]{%
amari1977}
\APACinsertmetastar {%
amari1977}%
\begin{APACrefauthors}%
Amari, S\BHBI i.%
\end{APACrefauthors}%
\unskip\
\newblock
\APACrefYearMonthDay{1977}{}{}.
\newblock
{\BBOQ}\APACrefatitle {Dynamics of pattern formation in lateral-inhibition type
  neural fields} {Dynamics of pattern formation in lateral-inhibition type
  neural fields}.{\BBCQ}
\newblock
\APACjournalVolNumPages{Biological Cybernetics}{27}{2}{77--87}.
\PrintBackRefs{\CurrentBib}

\bibitem [\protect \citeauthoryear {%
Babel%
}{%
Babel%
}{%
{\protect \APACyear {2010}}%
}]{%
babel2010}
\APACinsertmetastar {%
babel2010}%
\begin{APACrefauthors}%
Babel, M.%
\end{APACrefauthors}%
\unskip\
\newblock
\APACrefYearMonthDay{2010}{}{}.
\newblock
{\BBOQ}\APACrefatitle {Dialect divergence and convergence in {New Zealand
  English}} {Dialect divergence and convergence in {New Zealand
  English}}.{\BBCQ}
\newblock
\APACjournalVolNumPages{Language in Society}{39}{4}{437--456}.
\PrintBackRefs{\CurrentBib}

\bibitem [\protect \citeauthoryear {%
Babel%
}{%
Babel%
}{%
{\protect \APACyear {2012}}%
}]{%
babel2012}
\APACinsertmetastar {%
babel2012}%
\begin{APACrefauthors}%
Babel, M.%
\end{APACrefauthors}%
\unskip\
\newblock
\APACrefYearMonthDay{2012}{}{}.
\newblock
{\BBOQ}\APACrefatitle {Evidence for phonetic and social selectivity in
  spontaneous phonetic imitation} {Evidence for phonetic and social selectivity
  in spontaneous phonetic imitation}.{\BBCQ}
\newblock
\APACjournalVolNumPages{Journal of Phonetics}{40}{1}{177-189}.
\PrintBackRefs{\CurrentBib}

\bibitem [\protect \citeauthoryear {%
Barreda%
}{%
Barreda%
}{%
{\protect \APACyear {2021}}%
}]{%
barreda2021}
\APACinsertmetastar {%
barreda2021}%
\begin{APACrefauthors}%
Barreda, S.%
\end{APACrefauthors}%
\unskip\
\newblock
\APACrefYearMonthDay{2021}{}{}.
\newblock
{\BBOQ}\APACrefatitle {{Fast Track}: Fast (nearly) automatic formant-tracking
  using {Praat}} {{Fast Track}: Fast (nearly) automatic formant-tracking using
  {Praat}}.{\BBCQ}
\newblock
\APACjournalVolNumPages{Linguistics Vanguard}{7}{1}{20200051}.
\PrintBackRefs{\CurrentBib}

\bibitem [\protect \citeauthoryear {%
Erlhagen%
\ \BBA {} Sch\"{o}ner%
}{%
Erlhagen%
\ \BBA {} Sch\"{o}ner%
}{%
{\protect \APACyear {2002}}%
}]{%
erlhagen-schoener2002}
\APACinsertmetastar {%
erlhagen-schoener2002}%
\begin{APACrefauthors}%
Erlhagen, W.%
\BCBT {}\ \BBA {} Sch\"{o}ner, G.%
\end{APACrefauthors}%
\unskip\
\newblock
\APACrefYearMonthDay{2002}{}{}.
\newblock
{\BBOQ}\APACrefatitle {Dynamic field theory of movement preparation} {Dynamic
  field theory of movement preparation}.{\BBCQ}
\newblock
\APACjournalVolNumPages{Psychological Review}{109}{3}{545--572}.
\PrintBackRefs{\CurrentBib}

\bibitem [\protect \citeauthoryear {%
Evans%
\ \BBA {} Iverson%
}{%
Evans%
\ \BBA {} Iverson%
}{%
{\protect \APACyear {2007}}%
}]{%
evans-iverson2007}
\APACinsertmetastar {%
evans-iverson2007}%
\begin{APACrefauthors}%
Evans, B\BPBI G.%
\BCBT {}\ \BBA {} Iverson, P.%
\end{APACrefauthors}%
\unskip\
\newblock
\APACrefYearMonthDay{2007}{}{}.
\newblock
{\BBOQ}\APACrefatitle {Plasticity in vowel perception and production: A study
  of accent change in young adults} {Plasticity in vowel perception and
  production: A study of accent change in young adults}.{\BBCQ}
\newblock
\APACjournalVolNumPages{Journal of the Acoustical Society of
  America}{121}{6}{3814--3826}.
\PrintBackRefs{\CurrentBib}

\bibitem [\protect \citeauthoryear {%
Fowler%
}{%
Fowler%
}{%
{\protect \APACyear {1980}}%
}]{%
fowler1980}
\APACinsertmetastar {%
fowler1980}%
\begin{APACrefauthors}%
Fowler, C\BPBI A.%
\end{APACrefauthors}%
\unskip\
\newblock
\APACrefYearMonthDay{1980}{}{}.
\newblock
{\BBOQ}\APACrefatitle {Coarticulation and theories of extrinsic timing}
  {Coarticulation and theories of extrinsic timing}.{\BBCQ}
\newblock
\APACjournalVolNumPages{Journal of Phonetics}{8}{1}{113--133}.
\PrintBackRefs{\CurrentBib}

\bibitem [\protect \citeauthoryear {%
Fruehwald%
\ \BBA {} Barreda%
}{%
Fruehwald%
\ \BBA {} Barreda%
}{%
{\protect \APACyear {2023}}%
}]{%
fruehwald_fasttrackpy}
\APACinsertmetastar {%
fruehwald_fasttrackpy}%
\begin{APACrefauthors}%
Fruehwald, J.%
\BCBT {}\ \BBA {} Barreda, S.%
\end{APACrefauthors}%
\unskip\
\newblock
\APACrefYearMonthDay{2023}{}{}.
\newblock
\APACrefbtitle {{fasttrackpy}.} {{fasttrackpy}.}
\newblock
\APAChowpublished {Zenodo}.
\newblock
\begin{APACrefURL} \url{https://doi.org/10.5281/ZENODO.10212099}
  \end{APACrefURL}
\PrintBackRefs{\CurrentBib}

\bibitem [\protect \citeauthoryear {%
Gafos%
}{%
Gafos%
}{%
{\protect \APACyear {2006}}%
}]{%
gafos2006}
\APACinsertmetastar {%
gafos2006}%
\begin{APACrefauthors}%
Gafos, A\BPBI I.%
\end{APACrefauthors}%
\unskip\
\newblock
\APACrefYearMonthDay{2006}{}{}.
\newblock
{\BBOQ}\APACrefatitle {Dynamics in grammar} {Dynamics in grammar}.{\BBCQ}
\newblock
\BIn{} L.~Goldstein, D.~Whalen\BCBL {}\ \BBA {} C\BPBI T.~Best\ (\BEDS),
  \APACrefbtitle {Laboratory Phonology 8: Varieties of Phonological Competence}
  {Laboratory phonology 8: Varieties of phonological competence}\ (\BPGS\
  51--79).
\newblock
\APACaddressPublisher{Berlin}{Mouton de Gruyter}.
\PrintBackRefs{\CurrentBib}

\bibitem [\protect \citeauthoryear {%
Giles%
}{%
Giles%
}{%
{\protect \APACyear {1973}}%
}]{%
giles1973}
\APACinsertmetastar {%
giles1973}%
\begin{APACrefauthors}%
Giles, H.%
\end{APACrefauthors}%
\unskip\
\newblock
\APACrefYearMonthDay{1973}{}{}.
\newblock
{\BBOQ}\APACrefatitle {Accent mobility: A model and some data} {Accent
  mobility: A model and some data}.{\BBCQ}
\newblock
\APACjournalVolNumPages{Anthropological Linguistics}{15}{2}{87--105}.
\PrintBackRefs{\CurrentBib}

\bibitem [\protect \citeauthoryear {%
Goldinger%
}{%
Goldinger%
}{%
{\protect \APACyear {1998}}%
}]{%
goldinger1998}
\APACinsertmetastar {%
goldinger1998}%
\begin{APACrefauthors}%
Goldinger, S\BPBI D.%
\end{APACrefauthors}%
\unskip\
\newblock
\APACrefYearMonthDay{1998}{}{}.
\newblock
{\BBOQ}\APACrefatitle {Echoes of echoes? An episodic theory of lexical access}
  {Echoes of echoes? an episodic theory of lexical access}.{\BBCQ}
\newblock
\APACjournalVolNumPages{Psychological Review}{105}{2}{251--279}.
\PrintBackRefs{\CurrentBib}

\bibitem [\protect \citeauthoryear {%
Grossberg%
}{%
Grossberg%
}{%
{\protect \APACyear {1980}}%
}]{%
grossberg1980}
\APACinsertmetastar {%
grossberg1980}%
\begin{APACrefauthors}%
Grossberg, S.%
\end{APACrefauthors}%
\unskip\
\newblock
\APACrefYearMonthDay{1980}{}{}.
\newblock
{\BBOQ}\APACrefatitle {Biological competition: Decision rules, pattern
  formation, and oscillations} {Biological competition: Decision rules, pattern
  formation, and oscillations}.{\BBCQ}
\newblock
\APACjournalVolNumPages{Proceedings of the National Academy of
  Sciences}{77}{4}{2338--2342}.
\PrintBackRefs{\CurrentBib}

\bibitem [\protect \citeauthoryear {%
Gubian%
, Cronenberg%
\BCBL {}\ \BBA {} Harrington%
}{%
Gubian%
\ \protect \BOthers {.}}{%
{\protect \APACyear {2023}}%
}]{%
gubian-etal2023}
\APACinsertmetastar {%
gubian-etal2023}%
\begin{APACrefauthors}%
Gubian, M.%
, Cronenberg, J.%
\BCBL {}\ \BBA {} Harrington, J.%
\end{APACrefauthors}%
\unskip\
\newblock
\APACrefYearMonthDay{2023}{}{}.
\newblock
{\BBOQ}\APACrefatitle {Phonetic and phonological sound changes in an
  agent-based model} {Phonetic and phonological sound changes in an agent-based
  model}.{\BBCQ}
\newblock
\APACjournalVolNumPages{Speech Communication}{147}{}{93--115}.
\PrintBackRefs{\CurrentBib}

\bibitem [\protect \citeauthoryear {%
Haken%
}{%
Haken%
}{%
{\protect \APACyear {1977}}%
}]{%
haken1977}
\APACinsertmetastar {%
haken1977}%
\begin{APACrefauthors}%
Haken, H.%
\end{APACrefauthors}%
\unskip\
\newblock
\APACrefYear{1977}.
\newblock
\APACrefbtitle {Synergetics: An Introduction} {Synergetics: An introduction}.
\newblock
\APACaddressPublisher{Berlin}{Springer-Verlag}.
\PrintBackRefs{\CurrentBib}

\bibitem [\protect \citeauthoryear {%
Harper%
}{%
Harper%
}{%
{\protect \APACyear {2021}}%
}]{%
harper2021}
\APACinsertmetastar {%
harper2021}%
\begin{APACrefauthors}%
Harper, S\BPBI K.%
\end{APACrefauthors}%
\unskip\
\newblock
\APACrefYear{2021}.
\newblock
\APACrefbtitle {Individual differences in phonetic variability and phonological
  representation} {Individual differences in phonetic variability and
  phonological representation}.
\newblock
\BUPhD, University of Southern California, Los Angeles, CA.
\PrintBackRefs{\CurrentBib}

\bibitem [\protect \citeauthoryear {%
Harrington%
, Gubian%
, Stevens%
\BCBL {}\ \BBA {} Schiel%
}{%
Harrington%
\ \protect \BOthers {.}}{%
{\protect \APACyear {2019}}%
}]{%
harrington-etal2019}
\APACinsertmetastar {%
harrington-etal2019}%
\begin{APACrefauthors}%
Harrington, J.%
, Gubian, M.%
, Stevens, M.%
\BCBL {}\ \BBA {} Schiel, F.%
\end{APACrefauthors}%
\unskip\
\newblock
\APACrefYearMonthDay{2019}{}{}.
\newblock
{\BBOQ}\APACrefatitle {Phonetic change in an {A}ntarctic winter} {Phonetic
  change in an {A}ntarctic winter}.{\BBCQ}
\newblock
\APACjournalVolNumPages{Journal of the Acoustical Society of
  America}{146}{5}{3327--3332}.
\PrintBackRefs{\CurrentBib}

\bibitem [\protect \citeauthoryear {%
Harris%
\ \protect \BOthers {.}}{%
Harris%
\ \protect \BOthers {.}}{%
{\protect \APACyear {2020}}%
}]{%
harris2020}
\APACinsertmetastar {%
harris2020}%
\begin{APACrefauthors}%
Harris, C\BPBI R.%
, Millman, K\BPBI J.%
, van~der Walt, S\BPBI J.%
, Gommers, R.%
, Virtanen, P.%
, Cournapeau, D.%
\BDBL {}Oliphant, T\BPBI E.%
\end{APACrefauthors}%
\unskip\
\newblock
\APACrefYearMonthDay{2020}{}{}.
\newblock
{\BBOQ}\APACrefatitle {Array programming with {NumPy}} {Array programming with
  {NumPy}}.{\BBCQ}
\newblock
\APACjournalVolNumPages{Nature}{585}{7825}{357--362}.
\PrintBackRefs{\CurrentBib}

\bibitem [\protect \citeauthoryear {%
Hintzman%
}{%
Hintzman%
}{%
{\protect \APACyear {1984}}%
}]{%
hintzman1984}
\APACinsertmetastar {%
hintzman1984}%
\begin{APACrefauthors}%
Hintzman, D\BPBI L.%
\end{APACrefauthors}%
\unskip\
\newblock
\APACrefYearMonthDay{1984}{}{}.
\newblock
{\BBOQ}\APACrefatitle {Minerva 2: A simulation model of human memory} {Minerva
  2: A simulation model of human memory}.{\BBCQ}
\newblock
\APACjournalVolNumPages{Behavior Research Methods, Instruments, \&
  Computers}{16}{1}{96--101}.
\PrintBackRefs{\CurrentBib}

\bibitem [\protect \citeauthoryear {%
Houghton%
\ \BBA {} Tipper%
}{%
Houghton%
\ \BBA {} Tipper%
}{%
{\protect \APACyear {1996}}%
}]{%
houghton-tipper1996}
\APACinsertmetastar {%
houghton-tipper1996}%
\begin{APACrefauthors}%
Houghton, G.%
\BCBT {}\ \BBA {} Tipper, S\BPBI P.%
\end{APACrefauthors}%
\unskip\
\newblock
\APACrefYearMonthDay{1996}{}{}.
\newblock
{\BBOQ}\APACrefatitle {Inhibitory mechanisms of neural and cognitive control:
  Applications to selective attention and sequential action} {Inhibitory
  mechanisms of neural and cognitive control: Applications to selective
  attention and sequential action}.{\BBCQ}
\newblock
\APACjournalVolNumPages{Brain and Cognition}{30}{1}{20--43}.
\PrintBackRefs{\CurrentBib}

\bibitem [\protect \citeauthoryear {%
Hunter%
}{%
Hunter%
}{%
{\protect \APACyear {2007}}%
}]{%
hunter2007}
\APACinsertmetastar {%
hunter2007}%
\begin{APACrefauthors}%
Hunter, J\BPBI D.%
\end{APACrefauthors}%
\unskip\
\newblock
\APACrefYearMonthDay{2007}{}{}.
\newblock
{\BBOQ}\APACrefatitle {Matplotlib: A {2D} graphics environment} {Matplotlib: A
  {2D} graphics environment}.{\BBCQ}
\newblock
\APACjournalVolNumPages{Computing in Science \& Engineering}{9}{3}{90--95}.
\PrintBackRefs{\CurrentBib}

\bibitem [\protect \citeauthoryear {%
Johnson%
}{%
Johnson%
}{%
{\protect \APACyear {2007}}%
}]{%
johnson2007}
\APACinsertmetastar {%
johnson2007}%
\begin{APACrefauthors}%
Johnson, K.%
\end{APACrefauthors}%
\unskip\
\newblock
\APACrefYearMonthDay{2007}{}{}.
\newblock
{\BBOQ}\APACrefatitle {Decisions and mechanisms in Exemplar-based phonology}
  {Decisions and mechanisms in exemplar-based phonology}.{\BBCQ}
\newblock
\BIn{} M\BHBI J.~Sol\'e, P\BPBI S.~Beddor\BCBL {}\ \BBA {} M.~Ohala\ (\BEDS),
  \APACrefbtitle {Experimental Approaches to Phonology} {Experimental
  approaches to phonology}\ (\BPGS\ 25--40).
\newblock
\APACaddressPublisher{Oxford}{Oxford University Press}.
\PrintBackRefs{\CurrentBib}

\bibitem [\protect \citeauthoryear {%
Kelso%
}{%
Kelso%
}{%
{\protect \APACyear {1995}}%
}]{%
kelso1995}
\APACinsertmetastar {%
kelso1995}%
\begin{APACrefauthors}%
Kelso, J\BPBI S.%
\end{APACrefauthors}%
\unskip\
\newblock
\APACrefYear{1995}.
\newblock
\APACrefbtitle {Dynamic Patterns: The self-organization of brain and behavior}
  {Dynamic patterns: The self-organization of brain and behavior}.
\newblock
\APACaddressPublisher{Cambridge, MA}{MIT Press}.
\PrintBackRefs{\CurrentBib}

\bibitem [\protect \citeauthoryear {%
Kelso%
, Saltzman%
\BCBL {}\ \BBA {} Tuller%
}{%
Kelso%
\ \protect \BOthers {.}}{%
{\protect \APACyear {1986}}%
}]{%
kelso-etal1986}
\APACinsertmetastar {%
kelso-etal1986}%
\begin{APACrefauthors}%
Kelso, J\BPBI S.%
, Saltzman, E\BPBI L.%
\BCBL {}\ \BBA {} Tuller, B.%
\end{APACrefauthors}%
\unskip\
\newblock
\APACrefYearMonthDay{1986}{}{}.
\newblock
{\BBOQ}\APACrefatitle {The dynamical perspective on speech production: data and
  theory} {The dynamical perspective on speech production: data and
  theory}.{\BBCQ}
\newblock
\APACjournalVolNumPages{Journal of Phonetics}{14}{1}{29--59}.
\PrintBackRefs{\CurrentBib}

\bibitem [\protect \citeauthoryear {%
Kirkham%
}{%
Kirkham%
}{%
{\protect \APACyear {2025}}%
{\protect \APACexlab {{\protect \BCnt {1}}}}}]{%
kirkham2025b}
\APACinsertmetastar {%
kirkham2025b}%
\begin{APACrefauthors}%
Kirkham, S.%
\end{APACrefauthors}%
\unskip\
\newblock
\APACrefYearMonthDay{2025{\protect \BCnt {1}}}{}{}.
\newblock
{\BBOQ}\APACrefatitle {Discovering dynamical laws for speech gestures}
  {Discovering dynamical laws for speech gestures}.{\BBCQ}
\newblock
\APACjournalVolNumPages{Cognitive Science}{49}{5}{e70064}.
\PrintBackRefs{\CurrentBib}

\bibitem [\protect \citeauthoryear {%
Kirkham%
}{%
Kirkham%
}{%
{\protect \APACyear {2025}}%
{\protect \APACexlab {{\protect \BCnt {2}}}}}]{%
kirkham2025}
\APACinsertmetastar {%
kirkham2025}%
\begin{APACrefauthors}%
Kirkham, S.%
\end{APACrefauthors}%
\unskip\
\newblock
\APACrefYearMonthDay{2025{\protect \BCnt {2}}}{}{}.
\newblock
{\BBOQ}\APACrefatitle {Scaling laws for nonlinear dynamical models of
  articulatory control} {Scaling laws for nonlinear dynamical models of
  articulatory control}.{\BBCQ}
\newblock
\APACjournalVolNumPages{{JASA} Express Letters}{5}{2}{1--7}.
\PrintBackRefs{\CurrentBib}

\bibitem [\protect \citeauthoryear {%
Kirkham%
\ \BBA {} Strycharczuk%
}{%
Kirkham%
\ \BBA {} Strycharczuk%
}{%
{\protect \APACyear {2024}}%
}]{%
kirkham-strycharczuk2024}
\APACinsertmetastar {%
kirkham-strycharczuk2024}%
\begin{APACrefauthors}%
Kirkham, S.%
\BCBT {}\ \BBA {} Strycharczuk, P.%
\end{APACrefauthors}%
\unskip\
\newblock
\APACrefYearMonthDay{2024}{}{}.
\newblock
{\BBOQ}\APACrefatitle {A dynamic neural field model of vowel diphthongisation}
  {A dynamic neural field model of vowel diphthongisation}.{\BBCQ}
\newblock
\APACjournalVolNumPages{Proc. ISSP 2024 -- 13th International Seminar on Speech
  Production}{}{}{193--196}.
\PrintBackRefs{\CurrentBib}

\bibitem [\protect \citeauthoryear {%
McAuliffe%
, Socolof%
, Mihuc%
, Wagner%
\BCBL {}\ \BBA {} Sonderegger%
}{%
McAuliffe%
\ \protect \BOthers {.}}{%
{\protect \APACyear {2017}}%
}]{%
mcauliffe-etal2017}
\APACinsertmetastar {%
mcauliffe-etal2017}%
\begin{APACrefauthors}%
McAuliffe, M.%
, Socolof, M.%
, Mihuc, S.%
, Wagner, M.%
\BCBL {}\ \BBA {} Sonderegger, M.%
\end{APACrefauthors}%
\unskip\
\newblock
\APACrefYearMonthDay{2017}{}{}.
\newblock
{\BBOQ}\APACrefatitle {{Montreal Forced Aligner}: Trainable Text-Speech
  Alignment Using {Kaldi}} {{Montreal Forced Aligner}: Trainable text-speech
  alignment using {Kaldi}}.{\BBCQ}
\newblock
\BIn{} \APACrefbtitle {Proc. {Interspeech} 2017} {Proc. {Interspeech} 2017}\
  (\BPGS\ 498--502).
\PrintBackRefs{\CurrentBib}

\bibitem [\protect \citeauthoryear {%
Namy%
, Nygaard%
\BCBL {}\ \BBA {} Sauerteig%
}{%
Namy%
\ \protect \BOthers {.}}{%
{\protect \APACyear {2002}}%
}]{%
namy-etal2002}
\APACinsertmetastar {%
namy-etal2002}%
\begin{APACrefauthors}%
Namy, L\BPBI L.%
, Nygaard, L\BPBI C.%
\BCBL {}\ \BBA {} Sauerteig, D.%
\end{APACrefauthors}%
\unskip\
\newblock
\APACrefYearMonthDay{2002}{}{}.
\newblock
{\BBOQ}\APACrefatitle {Gender differences in vocal accommodation: The role of
  perception} {Gender differences in vocal accommodation: The role of
  perception}.{\BBCQ}
\newblock
\APACjournalVolNumPages{Journal of Language and Social
  Psychology}{21}{4}{422--432}.
\PrintBackRefs{\CurrentBib}

\bibitem [\protect \citeauthoryear {%
Pardo%
}{%
Pardo%
}{%
{\protect \APACyear {2006}}%
}]{%
pardo2006}
\APACinsertmetastar {%
pardo2006}%
\begin{APACrefauthors}%
Pardo, J\BPBI S.%
\end{APACrefauthors}%
\unskip\
\newblock
\APACrefYearMonthDay{2006}{}{}.
\newblock
{\BBOQ}\APACrefatitle {On phonetic convergence during conversational
  interaction} {On phonetic convergence during conversational
  interaction}.{\BBCQ}
\newblock
\APACjournalVolNumPages{Journal of the Acoustical Society of
  America}{119}{4}{2382--2393}.
\PrintBackRefs{\CurrentBib}

\bibitem [\protect \citeauthoryear {%
Peirce%
\ \protect \BOthers {.}}{%
Peirce%
\ \protect \BOthers {.}}{%
{\protect \APACyear {2019}}%
}]{%
psychopy2019}
\APACinsertmetastar {%
psychopy2019}%
\begin{APACrefauthors}%
Peirce, J.%
, Gray, J\BPBI R.%
, Simpson, S.%
, MacAskill, M.%
, H\"{o}chenberger, R.%
, Sogo, H.%
\BDBL {}Lindel\o{}v, J\BPBI K.%
\end{APACrefauthors}%
\unskip\
\newblock
\APACrefYearMonthDay{2019}{}{}.
\newblock
{\BBOQ}\APACrefatitle {{PsychoPy2}: experiments in behavior made easy}
  {{PsychoPy2}: experiments in behavior made easy}.{\BBCQ}
\newblock
\APACjournalVolNumPages{Behavior Research Methods}{51}{1}{195--203}.
\PrintBackRefs{\CurrentBib}

\bibitem [\protect \citeauthoryear {%
Pierrehumbert%
}{%
Pierrehumbert%
}{%
{\protect \APACyear {2002}}%
}]{%
pierrehumbert2002}
\APACinsertmetastar {%
pierrehumbert2002}%
\begin{APACrefauthors}%
Pierrehumbert, J\BPBI B.%
\end{APACrefauthors}%
\unskip\
\newblock
\APACrefYearMonthDay{2002}{}{}.
\newblock
{\BBOQ}\APACrefatitle {Word-specific phonetics} {Word-specific
  phonetics}.{\BBCQ}
\newblock
\BIn{} C.~Gussenhoven\ \BBA {} N.~Warner\ (\BEDS), \APACrefbtitle {Laboratory
  Phonology 7} {Laboratory phonology 7}\ (\BPGS\ 101--139).
\newblock
\APACaddressPublisher{Berlin}{Mouton de Gruyter}.
\PrintBackRefs{\CurrentBib}

\bibitem [\protect \citeauthoryear {%
Roon%
\ \BBA {} Gafos%
}{%
Roon%
\ \BBA {} Gafos%
}{%
{\protect \APACyear {2016}}%
}]{%
roon-gafos2016}
\APACinsertmetastar {%
roon-gafos2016}%
\begin{APACrefauthors}%
Roon, K\BPBI D.%
\BCBT {}\ \BBA {} Gafos, A\BPBI I.%
\end{APACrefauthors}%
\unskip\
\newblock
\APACrefYearMonthDay{2016}{}{}.
\newblock
{\BBOQ}\APACrefatitle {Perceiving while producing: Modeling the dynamics of
  phonological planning} {Perceiving while producing: Modeling the dynamics of
  phonological planning}.{\BBCQ}
\newblock
\APACjournalVolNumPages{Journal of Memory and Language}{89}{2}{222--243}.
\PrintBackRefs{\CurrentBib}

\bibitem [\protect \citeauthoryear {%
Samuelson%
, Smith%
, Perry%
\BCBL {}\ \BBA {} Spencer%
}{%
Samuelson%
\ \protect \BOthers {.}}{%
{\protect \APACyear {2011}}%
}]{%
samuelson-etal2011}
\APACinsertmetastar {%
samuelson-etal2011}%
\begin{APACrefauthors}%
Samuelson, L\BPBI K.%
, Smith, L\BPBI B.%
, Perry, L\BPBI K.%
\BCBL {}\ \BBA {} Spencer, J\BPBI P.%
\end{APACrefauthors}%
\unskip\
\newblock
\APACrefYearMonthDay{2011}{}{}.
\newblock
{\BBOQ}\APACrefatitle {Grounding word learning in space} {Grounding word
  learning in space}.{\BBCQ}
\newblock
\APACjournalVolNumPages{{PLoS ONE}}{6}{12}{e28095}.
\PrintBackRefs{\CurrentBib}

\bibitem [\protect \citeauthoryear {%
Sch\"{o}ner%
, Spencer%
\BCBL {}\ \BBA {} {The DFT Research Group}%
}{%
Sch\"{o}ner%
\ \protect \BOthers {.}}{%
{\protect \APACyear {2016}}%
}]{%
schoener-etal2016}
\APACinsertmetastar {%
schoener-etal2016}%
\begin{APACrefauthors}%
Sch\"{o}ner, G.%
, Spencer, J\BPBI P.%
\BCBL {}\ \BBA {} {The DFT Research Group}.%
\end{APACrefauthors}%
\unskip\
\newblock
\APACrefYear{2016}.
\newblock
\APACrefbtitle {Dynamic Thinking: A Primer on Dynamic Field Theory} {Dynamic
  thinking: A primer on dynamic field theory}.
\newblock
\APACaddressPublisher{Oxford}{Oxford University Press}.
\PrintBackRefs{\CurrentBib}

\bibitem [\protect \citeauthoryear {%
Shockey%
, Sabadini%
\BCBL {}\ \BBA {} Fowler%
}{%
Shockey%
\ \protect \BOthers {.}}{%
{\protect \APACyear {2004}}%
}]{%
shockley-etal2004}
\APACinsertmetastar {%
shockley-etal2004}%
\begin{APACrefauthors}%
Shockey, K.%
, Sabadini, L.%
\BCBL {}\ \BBA {} Fowler, C\BPBI A.%
\end{APACrefauthors}%
\unskip\
\newblock
\APACrefYearMonthDay{2004}{}{}.
\newblock
{\BBOQ}\APACrefatitle {Imitation in shadowing words} {Imitation in shadowing
  words}.{\BBCQ}
\newblock
\APACjournalVolNumPages{Perception \& Psychophysics}{66}{3}{422--429}.
\PrintBackRefs{\CurrentBib}

\bibitem [\protect \citeauthoryear {%
Sorensen%
\ \BBA {} Gafos%
}{%
Sorensen%
\ \BBA {} Gafos%
}{%
{\protect \APACyear {2016}}%
}]{%
sorensen-gafos2016}
\APACinsertmetastar {%
sorensen-gafos2016}%
\begin{APACrefauthors}%
Sorensen, T.%
\BCBT {}\ \BBA {} Gafos, A\BPBI I.%
\end{APACrefauthors}%
\unskip\
\newblock
\APACrefYearMonthDay{2016}{}{}.
\newblock
{\BBOQ}\APACrefatitle {The gesture as an autonomous nonlinear dynamical system}
  {The gesture as an autonomous nonlinear dynamical system}.{\BBCQ}
\newblock
\APACjournalVolNumPages{Ecological Psychology}{28}{4}{188--215}.
\PrintBackRefs{\CurrentBib}

\bibitem [\protect \citeauthoryear {%
{Stan Development Team}%
}{%
{Stan Development Team}%
}{%
{\protect \APACyear {2024}}%
}]{%
stan2024}
\APACinsertmetastar {%
stan2024}%
\begin{APACrefauthors}%
{Stan Development Team}.%
\end{APACrefauthors}%
\unskip\
\newblock
\APACrefYearMonthDay{2024}{}{}.
\newblock
\APACrefbtitle {{Stan Reference Manual}, v2.36.0.} {{Stan Reference Manual},
  v2.36.0.}
\newblock
\APAChowpublished {https://mc-stan.org}.
\PrintBackRefs{\CurrentBib}

\bibitem [\protect \citeauthoryear {%
Stern%
\ \BBA {} Shaw%
}{%
Stern%
\ \BBA {} Shaw%
}{%
{\protect \APACyear {2023}}%
}]{%
stern-shaw2023}
\APACinsertmetastar {%
stern-shaw2023}%
\begin{APACrefauthors}%
Stern, M\BPBI C.%
\BCBT {}\ \BBA {} Shaw, J\BPBI A.%
\end{APACrefauthors}%
\unskip\
\newblock
\APACrefYearMonthDay{2023}{}{}.
\newblock
{\BBOQ}\APACrefatitle {Neural inhibition during speech planning contributes to
  contrastive hyperarticulation} {Neural inhibition during speech planning
  contributes to contrastive hyperarticulation}.{\BBCQ}
\newblock
\APACjournalVolNumPages{Journal of Memory and Language}{132}{104443}{1--16}.
\PrintBackRefs{\CurrentBib}

\bibitem [\protect \citeauthoryear {%
Stern%
\ \BBA {} Shaw%
}{%
Stern%
\ \BBA {} Shaw%
}{%
{\protect \APACyear {2024}}%
}]{%
stern-shaw2024}
\APACinsertmetastar {%
stern-shaw2024}%
\begin{APACrefauthors}%
Stern, M\BPBI C.%
\BCBT {}\ \BBA {} Shaw, J\BPBI A.%
\end{APACrefauthors}%
\unskip\
\newblock
\APACrefYearMonthDay{2024}{}{}.
\newblock
{\BBOQ}\APACrefatitle {Towards a minimal dynamics for gestures: A law relating
  velocity and position} {Towards a minimal dynamics for gestures: A law
  relating velocity and position}.{\BBCQ}
\newblock
\APACjournalVolNumPages{Proc. {ISSP} 2024 -- 13th International Seminar on
  Speech Production}{}{}{262--265}.
\PrintBackRefs{\CurrentBib}

\bibitem [\protect \citeauthoryear {%
Strycharczuk%
, Kirkham%
, Gorman%
\BCBL {}\ \BBA {} Nagamine%
}{%
Strycharczuk%
\ \protect \BOthers {.}}{%
{\protect \APACyear {2024}}%
}]{%
strycharczuk-etal2024}
\APACinsertmetastar {%
strycharczuk-etal2024}%
\begin{APACrefauthors}%
Strycharczuk, P.%
, Kirkham, S.%
, Gorman, E.%
\BCBL {}\ \BBA {} Nagamine, T.%
\end{APACrefauthors}%
\unskip\
\newblock
\APACrefYearMonthDay{2024}{}{}.
\newblock
{\BBOQ}\APACrefatitle {Towards a dynamical model of {E}nglish vowels: Evidence
  from diphthongisation} {Towards a dynamical model of {E}nglish vowels:
  Evidence from diphthongisation}.{\BBCQ}
\newblock
\APACjournalVolNumPages{Journal of Phonetics}{107}{}{1--26}.
\PrintBackRefs{\CurrentBib}

\bibitem [\protect \citeauthoryear {%
{The pandas development team}%
}{%
{The pandas development team}%
}{%
{\protect \APACyear {2020}}%
}]{%
pandas}
\APACinsertmetastar {%
pandas}%
\begin{APACrefauthors}%
{The pandas development team}.%
\end{APACrefauthors}%
\unskip\
\newblock
\APACrefYearMonthDay{2020}{}{}.
\newblock
\APACrefbtitle {{pandas-dev/pandas: Pandas}.} {{pandas-dev/pandas: Pandas}.}
\newblock
\APAChowpublished {https://doi.org/10.5281/zenodo.3509134}.
\PrintBackRefs{\CurrentBib}

\bibitem [\protect \citeauthoryear {%
{The plotnine development team}%
}{%
{The plotnine development team}%
}{%
{\protect \APACyear {2025}}%
}]{%
plotnine}
\APACinsertmetastar {%
plotnine}%
\begin{APACrefauthors}%
{The plotnine development team}.%
\end{APACrefauthors}%
\unskip\
\newblock
\APACrefYearMonthDay{2025}{}{}.
\newblock
\APACrefbtitle {{plotnine: A grammar of graphics for Python}.} {{plotnine: A
  grammar of graphics for Python}.}
\newblock
\APAChowpublished {https://doi.org/10.5281/zenodo.1325308}.
\PrintBackRefs{\CurrentBib}

\bibitem [\protect \citeauthoryear {%
Tilsen%
}{%
Tilsen%
}{%
{\protect \APACyear {2007}}%
}]{%
tilsen2007}
\APACinsertmetastar {%
tilsen2007}%
\begin{APACrefauthors}%
Tilsen, S.%
\end{APACrefauthors}%
\unskip\
\newblock
\APACrefYearMonthDay{2007}{}{}.
\newblock
{\BBOQ}\APACrefatitle {Vowel-to-vowel coarticulation and dissimilation in
  phonemic-response priming} {Vowel-to-vowel coarticulation and dissimilation
  in phonemic-response priming}.{\BBCQ}
\newblock
\APACjournalVolNumPages{UC Berkeley Phonology Lab Annual
  Report}{3}{1}{416--458}.
\PrintBackRefs{\CurrentBib}

\bibitem [\protect \citeauthoryear {%
Tilsen%
}{%
Tilsen%
}{%
{\protect \APACyear {2009}}%
}]{%
tilsen2009b}
\APACinsertmetastar {%
tilsen2009b}%
\begin{APACrefauthors}%
Tilsen, S.%
\end{APACrefauthors}%
\unskip\
\newblock
\APACrefYearMonthDay{2009}{}{}.
\newblock
{\BBOQ}\APACrefatitle {Subphonemic and cross-phonemic priming in vowel
  shadowing: Evidence for the involvement of exemplars in production}
  {Subphonemic and cross-phonemic priming in vowel shadowing: Evidence for the
  involvement of exemplars in production}.{\BBCQ}
\newblock
\APACjournalVolNumPages{Journal of Phonetics}{37}{3}{276--296}.
\PrintBackRefs{\CurrentBib}

\bibitem [\protect \citeauthoryear {%
Tilsen%
}{%
Tilsen%
}{%
{\protect \APACyear {2019}}%
}]{%
tilsen2019}
\APACinsertmetastar {%
tilsen2019}%
\begin{APACrefauthors}%
Tilsen, S.%
\end{APACrefauthors}%
\unskip\
\newblock
\APACrefYearMonthDay{2019}{}{}.
\newblock
{\BBOQ}\APACrefatitle {Motoric mechanisms for the emergence of non-local
  phonological patterns} {Motoric mechanisms for the emergence of non-local
  phonological patterns}.{\BBCQ}
\newblock
\APACjournalVolNumPages{Frontiers in Psychology}{10}{2143}{1--25}.
\PrintBackRefs{\CurrentBib}

\bibitem [\protect \citeauthoryear {%
Virtanen%
\ \protect \BOthers {.}}{%
Virtanen%
\ \protect \BOthers {.}}{%
{\protect \APACyear {2020}}%
}]{%
SciPy-NMeth2020}
\APACinsertmetastar {%
SciPy-NMeth2020}%
\begin{APACrefauthors}%
Virtanen, P.%
, Gommers, R.%
, Oliphant, T\BPBI E.%
, Haberland, M.%
, Reddy, T.%
, Cournapeau, D.%
\BDBL {}{SciPy 1.0 Contributors}%
\end{APACrefauthors}%
\unskip\
\newblock
\APACrefYearMonthDay{2020}{}{}.
\newblock
{\BBOQ}\APACrefatitle {{{SciPy} 1.0: Fundamental Algorithms for Scientific
  Computing in Python}} {{{SciPy} 1.0: Fundamental Algorithms for Scientific
  Computing in Python}}.{\BBCQ}
\newblock
\APACjournalVolNumPages{Nature Methods}{17}{}{261--272}.
\PrintBackRefs{\CurrentBib}

\bibitem [\protect \citeauthoryear {%
Wells%
}{%
Wells%
}{%
{\protect \APACyear {1982}}%
}]{%
wells1982}
\APACinsertmetastar {%
wells1982}%
\begin{APACrefauthors}%
Wells, J\BPBI C.%
\end{APACrefauthors}%
\unskip\
\newblock
\APACrefYear{1982}.
\newblock
\APACrefbtitle {Accents of {E}nglish: Volumes 1--3} {Accents of {E}nglish:
  Volumes 1--3}.
\newblock
\APACaddressPublisher{Cambridge}{Cambridge University Press}.
\PrintBackRefs{\CurrentBib}

\end{thebibliography}

\end{document}